\documentclass{llncs}

\usepackage{helvet,times,courier}
\usepackage{microtype} 
\usepackage{graphicx}
\usepackage{amssymb}
\usepackage{amsmath}
\usepackage{mathnotation}
\usepackage{color}
\usepackage[ruled,vlined,linesnumbered]{algorithm2e}
\usepackage[draft]{fixme}
\usepackage{tikz}
\usepackage{pgfplots}
\usepackage{thmtools}
\usepackage{thm-restate}
\usepackage{caption}
\usepackage{subcaption}
\usepackage{url}

\usepackage{tabularx}
\usepackage{longtable}
\usepackage{multirow}

\fxsetup{
inline,
nomargin,
theme=color,
mode=multiuser
}
\FXRegisterAuthor{ks}{aks}{\color{blue}KS}
\FXRegisterAuthor{fr}{afr}{\color{green}FR}
\FXRegisterAuthor{cd}{acd}{\color{magenta}CD}
\FXRegisterAuthor{pg}{apg}{\color{orange}PG}

\SetFuncSty{textrm}
\SetKwFunction{GetCore}{ComputeUnsatCore}
\SetKwFunction{MinimizeCore}{MinimizeCore}
\SetKwFunction{GetQuery}{DetermineQuery}
\SetKwFunction{GetDiagnoses}{ComputeDiagnoses}
\SetKwInput{Input}{input}

\pgfplotsset{
  filter discard warning=false 
, legend cell align=left
, minor grid style={loosely dotted, lightgray}
, major grid style={loosely dashed, lightgray}
}
\pgfplotscreateplotcyclelist{cactus}{%
  red,      mark size=2.0pt, mark=star\\%
  blue,        mark size=2.0pt, mark=*\\%
  darkgreen,       mark size=2.0pt, mark=diamond*\\%
  black!80, mark size=2.0pt, mark=triangle*\\%
  gray,    mark size=2.0pt, mark=square\\%
}

\newcommand{\clasp}{\textsc{clasp}\xspace}
\newcommand{\wasp}{\textsc{wasp}\xspace}
\newcommand{\gringo}{\textsc{gringo}\xspace}
\newcommand{\minisat}{\textsc{minisat}\xspace}
\newcommand{\glucose}{\textsc{glucose}\xspace}
\newcommand{\lingeling}{\textsc{lingeling}\xspace}

\newcommand{\vsids}{\textsc{vsids}\xspace}

\newcommand{\inco}{\textsc{inconsistent}\xspace}
\newcommand{\consistent}{\textsc{consistent}\xspace}

\newcommand{\event}[2]{\textsc{On#1(#2)}\xspace}
\newcommand{\response}[2]{\textsc{\##1(#2)}\xspace}
\newcommand{\request}[2]{\textsc{#1}(#2)\xspace}

\newcommand{\pup}{PUP\xspace}
\newcommand{\ucap}{\textsc{UCap}\xspace}
\newcommand{\iucap}{\textsc{IUCap}\xspace}
\newcommand{\qpup}{\textsc{Quick\pup}\xspace}
\newcommand{\qspup}{$\textsc{Quick\pup}^*$\xspace}
\newcommand{\pred}{\textsc{Pred}\xspace}

\newcommand{\ccp}{CCP\xspace}
\newcommand{\accp}[1]{A#1\xspace}
\newcommand{\mccp}{A1A2\xspace}
\newcommand{\bccp}{A2F\xspace}
\newcommand{\bfccp}{A2FO\xspace}
\newcommand{\bfaccp}{A2AFO\xspace}

\definecolor{darkgreen}{RGB}{68,180,46}
\definecolor{darkgray}{RGB}{80,80,80}

\def\naf{\ensuremath{\raise.17ex\hbox{\ensuremath{\scriptstyle\mathtt{\sim}}}}\xspace}
\def\A{\ensuremath{\mathcal{A}}\xspace}
\def\L{\ensuremath{\mathcal{L}}\xspace}
\def\I{\ensuremath{\mathcal{I}}\xspace}
\def\F{\ensuremath{\mathcal{F}}\xspace}
\def\S{\ensuremath{\mathcal{S}}\xspace}

\title{Driving CDCL Search
}
\author{
	Carmine Dodaro\inst{1}%
	\and Philip Gasteiger\inst{2}%
	\and Nicola Leone\inst{1}%
	\and \\Benjamin Musitsch\inst{2}%
	\and Francesco Ricca\inst{1}%
	\and Konstantin Schekotihin\inst{2}%
}

\institute{
		Universit\`a della Calabria, Rende CS, IT\\%
		\email{\{lastname\}@mat.unical.it}
		\and  Alpen-Adria-Universit\"at Klagenfurt, 9020 Klagenfurt, AT\\%
		\email{\{firstname.lastname\}@aau.at}%
}

\begin{document}

\maketitle

\begin{abstract}
The CDCL algorithm is the leading solution adopted by state-of-the-art solvers for SAT, SMT, ASP, and others.
Experiments show that the performance of CDCL solvers can be significantly boosted by embedding domain-specific heuristics, especially on large real-world problems.
However, a proper integration of such criteria in off-the-shelf CDCL implementations is not obvious.
In this paper, we distill the key ingredients that drive the search of CDCL solvers, 
and propose a general framework for designing and implementing new heuristics.
We implemented our strategy in an ASP solver, and we experimented on two industrial domains.
On hard problem instances, state-of-the-art implementations fail to find any solution in acceptable time, whereas our implementation is very successful and finds all solutions.
\end{abstract}

\section{Introduction}
The Conflict Driven Clause Learning (CDCL) algorithm~\cite{DBLP:journals/tc/Marques-SilvaS99} is the leading solution adopted by state-of-the-art solvers for Boolean Satisfiability (SAT)~\cite{DBLP:series/faia/2009-185}, Satisfiability Modulo Theories (SMT)~\cite{DBLP:journals/jacm/NieuwenhuisOT06}, and Answer Set Programming (ASP)~\cite{DBLP:journals/cacm/BrewkaET11} to mention a few.
Notably, CDCL solvers have been applied with success for solving several real-world problems ranging from hardware and software model checking, planning, equivalence checking, bioinformatics, configuration problems, hardware and software test, software package dependencies, cryptography and more~\cite{DBLP:series/faia/2009-185}. 

As a matter of fact the performance of a CDCL solver heavily depends on the adoption of heuristics that drive the search for solutions. 
Among these, the heuristic for the selection of the branching literal 
(i.e., the criterion determining the literal to be assumed true at a given stage of the computation) 
can dramatically affect the overall performance of an implementation~\cite{DBLP:series/faia/2009-185}.
State-of-the-art CDCL implementations feature very good general purpose heuristics belonging to the family of VSIDS~\cite{DBLP:conf/sat/BiereF15,DBLP:conf/sat/EenS03,DBLP:conf/dac/MoskewiczMZZM01}. 
Since their introduction, VSIDS heuristics proved to be a key ingredient for solving many relevant problems~\cite{DBLP:conf/sat/BiereF15}.
Nonetheless, no general heuristic is known to be the best possible choice for all problems~\cite{DBLP:journals/ai/HutterXHL14}.
Pioneering work on employing domain-specific heuristics corroborates the validity of that idea. 
In particular, Beame et al. demonstrated the utility of a domain-specific branching heuristic for solving the pebbling formulas with a CDCL solver~\cite{DBLP:journals/jair/BeameKS04}; 
and Rintanen proposed to replace the standard VSIDS by a domain-heuristics for efficiently solving planning as satisfiability~\cite{DBLP:journals/ai/Rintanen12}.
Moreover, Gerhard Friedrich in his joint invited talk at CP-ICLP 2015 described a number of experiences in which ``domain-specific heuristics turned out to be 
the key component in several industrial applications of problem solvers''~\cite{InvitedFriedrich15}.
However, as argued in~\cite{DBLP:journals/ai/Rintanen12}: ``the main challenge in defining a variable selection scheme is its integration in the CDCL algorithm in a productive way''. 
Indeed, domain experts can quickly provide promising heuristic-criteria based on properties of solutions, but they may not have the knowledge and the experience to plug them in CDCL implementations.

The goal of the paper is to ease the design and the evaluation of new heuristics for CDCL solvers as well as to reduce the efforts needed to provide effective implementations.
To this end we distill the key ingredients that drive the search of CDCL solvers, 
and propose a general framework for designing and implementing new heuristics.
Useless to say that one could invent new heuristics based on other different ``ingredients''.
Nevertheless, our framework is quite rich and flexible, and indeed it covers the definition of the most popular heuristics for CDCL, as well as enables the definition of interesting problem-specific heuristics.
Moreover, further features could be easily incorporated.
A number of examples are provided to demonstrate the applicability of the proposed framework for defining a variety of known heuristic criteria from the literature~\cite{DBLP:conf/sat/BiereF15,DBLP:journals/ai/Rintanen12}. 
Pragmatic evidence of the benefits of our framework in devising problem-specific heuristics is provided by reporting on a use case where two hard industrial domains defined by Siemens, the Partner Units Problem (\pup)~\cite{DBLP:conf/iaai/TeppanFF12} and Combined Configuration Problem (\ccp)~\cite{DBLP:conf/lpnmr/GebserRS15}, are solved efficiently.

In our use case, the framework has been implemented by extending the ASP solver \wasp~\cite{DBLP:conf/lpnmr/AlvianoDLR15}.
Our implementation offers a wide range of possibilities to end users for developing new heuristics.
In particular, it offers multi-language support including scripting languages for fast prototyping, an embedded C++ interface for performance-oriented implementations, as well as heuristic definition by means of declarative predicate-based paradigms. 

Experiments on \pup and \ccp confirm the viability of the framework in real-world scenarios, and the effectiveness of our implementation that is able to solve successfully hard problem instances which were not solvable by state-of-the-art ASP solvers. Furthermore, we were able to solve hardest \ccp instances for which no solution was known before.

\section{An overview of the CDCL algorithm}
In this section we provide some basic knowledge about the Conflict-Driven Clause Learning (CDCL) algorithm~\cite{DBLP:journals/tc/Marques-SilvaS99} for solving the satisfiability problem (SAT)
~\cite{DBLP:conf/sat/BiereF15}. %
The Conflict-Driven Clause Learning (CDCL) \cite{DBLP:journals/tc/Marques-SilvaS99} backtracking search algorithm is a de facto standard for the satisfiability problem (SAT)~\cite{DBLP:conf/ijcai/AudemardS09,DBLP:conf/sat/BiereF15}.
During the recent years, CDCL has been extended in order to fit specific requirements of formalisms in the neighborhood of SAT, e.g. Satisfiability Modulo Theories and Answer Set Programming.
In the following, we first introduce some basic concepts of the satisfiability problem and then we describe the CDCL algorithm.

\paragraph{SAT.}
Let $\A$ be a fixed, countable set of propositional atoms including $\bot$. 
A literal $\ell$ is either an atom $a$ or its negation $\neg a$,
and $atom(\ell)=a$ denotes the atom associated to $\ell$.
The complement of $\ell$ is denoted by $\overline{\ell}$, where $\overline{a} = \neg a$ and $\overline{\neg a} = a$ for an atom $a$.
For a set $L$ of literals, $\overline{L} := \{\overline{\ell} \mid \ell \in L\}$, $L^+ :=L \cap \A$, and $L^- := \overline{L} \cap \A$.
A (partial) interpretation is a set of literals $I$ containing $\lnot \bot$. 
$I$ is inconsistent if $I^+ \cap I^- \neq \emptyset$, otherwise $I$ is consistent.
$I$ is total if $I^+ \cup I^- = \A$.
Given an interpretation $I$, a literal $\ell$ is true if $\ell \in I$; is false if $\overline{\ell} \in I$, and is undefined otherwise. 
A clause $\varphi$ is a set of literals, and a formula $\Gamma$ is a set of clauses.
An interpretation $I$ satisfies a clause $\varphi$ if $\varphi \cap I \neq \emptyset$.
A consistent, total interpretation $I$ is a model of a formula $\Gamma$ if all clauses in $\Gamma$ are satisfied.
A formula $\Gamma$ is said to be consistent if it admits some model, inconsistent otherwise.
Given a formula $\Gamma$, the SAT problem consists in determining whether $\Gamma$ is consistent.
\begin{example}\label{ex:ex1}
	Consider the formula $\Gamma=\{\{a, b,\neg c\}, \{a\},$ $ \{\neg b\}, \{c, d\}, \{c, \neg d\}\}$.
	$I=\{a,\neg b, c, d\}$ is a model of $\Gamma$.
\end{example}

\paragraph{CDCL algorithm.}
The CDCL algorithm takes as input a formula $\Gamma$, and determines whether it is consistent.
The first step of the algorithm consists of the simplification of $\Gamma$. 
Polynomial algorithms for clause rewriting and variable elimination are applied on $\Gamma$, which strengthen and/or remove redundant clauses \cite{DBLP:conf/sat/EenB05}.
After the simplifications step, the partial interpretation $I$ is set to $\{\lnot \bot\}$, and the backtracking search starts.
First, $I$ is extended with all the literals that can be deterministically inferred by applying some inference rule (\textit{propagation} step). The main inference rule of a SAT solver is \textit{unit propagation} that extends $I$ by an undefined literal $\ell$ whenever there is a clause $\varphi\in\Gamma$ such that $\ell \in \varphi$ and $\varphi \setminus \ell \subseteq \overline{I}$.
Three cases are possible after a propagation step is completed:
$(i)$ $I$ is consistent but not total. In that case, an undefined literal $\ell$ (called \textit{branching literal}) is chosen according to some heuristic criterion, and is added to $I$.
Then, a propagation step is performed that infers the consequences of this choice.
$(ii)$ $I$ is inconsistent, thus there is a conflict, and $I$ is analyzed. The reason of the conflict is modeled by a fresh clause $\varphi$ that is added to $\Gamma$ (clause \textit{learning}), e.g. by applying the \emph{first Unique Implication Point (UIP)} technique \cite{DBLP:conf/iccad/ZhangMMM01}.
Moreover, the algorithm backtracks (i.e. choices and their consequences are undone) until the consistency of $I$ is restored.
Then the algorithm propagates inferences starting from the fresh clause $\varphi$. Otherwise, if the consistency of $I$ cannot be restored, the algorithm terminates returning \inco.
Finally, in case $(iii)$ $I$ is consistent and total, i.e. $I$ is a model, the algorithm terminates returning \consistent.

\begin{example}\label{ex:ex2}
	Consider the formula $\Gamma$ of Example~\ref{ex:ex1}. 
	The search starts with $I:=\{\lnot \bot\}$.
	Unit propagation extends $I$ with $a$ and $\neg b$ to satisfy clauses $\{a\}$ and $\{\neg b\}$.
	No other inferences can be done, thus a choice is performed among $c$, $\neg c$, $d$, and $\neg d$.
	Choosing $\neg c$ would lead to an inconsistency, 
	while all other choices would directly lead to a model.
\end{example}

\paragraph{Heuristics.} It is well-known that the selection the branching literal plays a crucial role for efficiency.
Nowadays, state-of-the-art solvers, such as \lingeling~\cite{DBLP:conf/sat/BiereF15} and \glucose~\cite{DBLP:conf/ijcai/AudemardS09}, implement branching criteria belonging to the VSIDS~\cite{DBLP:conf/dac/MoskewiczMZZM01} family that are a variant of the \minisat~\cite{DBLP:conf/sat/EenS03} heuristic.
Thus, in the following, the \minisat~\cite{DBLP:conf/sat/EenS03} heuristic is referred to as the \textit{default branching heuristic} of a CDCL solver.
The \minisat heuristic is based on the \textit{activity} value for each atom in the input formula $\Gamma$, which is initially set to 0. The activity of $atom(\ell)$ is incremented by a value $inc$ when a literal $\ell$ occurs in a learned clause. Then, after learning the clause, the value of $inc$ is multiplied by a constant slightly greater than 1, to give more and more importance to variables that occur in recently-learned clauses.
Once a choice is needed, the literal $\lnot at$ is chosen, where $at$ is the undefined atom having the highest activity value (ties are broken randomly).

For the sake of completeness, we mention that the CDCL is usually complemented with heuristics that control the number of learned clauses, and restart the computation to explore different branches of the search tree~\cite{DBLP:series/faia/2009-185}.

\section{Driving CDCL Search}
\label{sec:heur}

\begin{table}[t!]
	\centering	
	\begin{tabular}{|p{.9\textwidth}|}
		\multicolumn{1}{c}{\textbf{CDCL $\longrightarrow$  Driver: Events} } \\[0.4em]
		\hline
		\\[-0.8em]
		\multicolumn{1}{|c|}{\textbf{\event{Search}{$\Gamma',\A'$}}} \\
		\textit{Triggered when the backtracking search starts. $\Gamma'$ and $\A'$ are the formula and the atoms after the simplifications, resp.}\\
		\hline
		\hline
		\\[-0.8em]
		\multicolumn{1}{|c|}{\textbf{\event{IncoChoice}{$\ell$}}}\\
		\textit{Triggered when the choice $\ell$ led to an inconsistency.}\\
		\hline \hline
		\\[-0.8em]
		\multicolumn{1}{|c|}{\textbf{\event{Conflict}{$\ell$}}}\\
		\textit{Triggered when a conflict is detected after the choice of $\ell$.}\\
		\hline \hline
		\\[-0.8em]
		\multicolumn{1}{|c|}{\textbf{\event{LearnClause}{$\varphi$}}}\\
		\textit{Triggered when the clause $\varphi$ is learned.} \\
		\hline \hline
		\\[-0.8em]
		\multicolumn{1}{|c|}{\textbf{\event{LitInConflict}{$\ell$}}}\\
		\textit{Triggered when the literal $\ell$ is involved in the creation of a clause to learn, e.g., while applying the first UIP schema.}\\
		\hline \hline
		\\[-0.8em]
		\multicolumn{1}{|c|}{\textbf{\event{Deletion}{$\varphi$}}}\\
		\textit{Triggered when the clause $\varphi$ has been removed.}\\		
		\hline \hline
		\\[-0.8em]
		\multicolumn{1}{|c|}{\textbf{\event{Restart}}} \\
		\textit{Triggered when a restart of the search occurs.}\\
		\hline \hline
		\\[-0.8em]
		\multicolumn{1}{|c|}{\textbf{\event{UnrollLit}{$\ell$}}} \\
		\textit{Triggered during backtracking if the literal $\ell$ is set to undefined.}\\
		\hline
	\end{tabular}
	\caption{Events notified by the CDCL solver.\label{tab:events}}
\end{table}

In this section we present the results of our studies pinpointing the key ingredients needed to drive the search of CDCL solvers. 
These were distilled by considering the needs of well-known heuristics, as well as by applying the lessons learned while developing domain-specific heuristics in the case study reported in the next section.

In our framework we identify two main actors: the \textit{solver} and the \textit{driver}. 
The first is, basically, an implementation of the CDCL algorithm, and the second implements an external branching heuristic.

Selection of a branching literal may depend (or not depend) on properties of the input, on the (current) solver state, on employed optimization techniques (e.g., input rewritings), as well as on the status of variables that directly model some properties of the problem, etc. 
This information is possibly communicated by the solver to the driver by means of \textit{events}. 
Events can be seen as asynchronous messages sent to the driver when specific points of the computation are reached, as detailed in Table~\ref{tab:events}.
The driver could exploit this information to implement the heuristic criteria, e.g., updating counters or modifying its status. None of the events are mandatory for the driver (we expect that an efficient implementation smartly sends only those required by the driver).

We have also identified two main moments in the computation when the solver gets directions from the driver. These are modeled in our framework by means of \textit{requests}, that are detailed in Table~\ref{tab:responses} together with the \textit{responses} the solver expects to receive.
Requests are synchronous, i.e., the solver waits for a response of (i.e., a command from) the driver.

The request \request{GetAtomsToBeFrozen}{$\A$} is sent before starting the simplifications of the input formula. 
In fact, preprocessing can remove atoms that the heuristic strategy may want to keep (they may play a role in a domain-specific heuristic). The response \response{Freeze}{$A \subseteq \A$} informs the solver to avoid eliminating the variables in $\A$.

\begin{table}[t!]
	\centering
	\begin{tabular}{|p{.9\textwidth}|}
		\multicolumn{1}{c}{\textbf{CDCL $\longleftrightarrow$  Driver: Requests \& Responses}}\\[0.4em]
		\hline
		\\[-0.8em]
		\multicolumn{1}{|c|}{\textbf{\request{GetAtomsToBeFrozen}{$\A$}}} \\
		\textit{Asks the heuristic strategy to provide a list of atoms $A \subseteq \A$ that must not be removed by simplifications.}\vspace*{3pt}\\
		\multicolumn{1}{|c|}{Allowed Response:} \\
		\begin{minipage}[c]{.89\textwidth}
			\begin{tabular}{|p{\textwidth}|}
				\hline
				\\[-0.8em]
				\multicolumn{1}{|c|}{\textbf{\response{Freeze}{$A$}}} \\
				\textit{Asks to not remove atoms in $A$ during preprocessing.}\\
				\hline
			\end{tabular}	
		\end{minipage}\vspace*{2pt}\\
		\hline
		\hline
		\\[-0.8em]
		\multicolumn{1}{|c|}{\textbf{\request{GetChoice}{$I$}}}\\
		\textit{Asks the heuristic strategy to provide the next branching literal(s) given the current interpretation $I$.} \vspace*{3pt}\\
		\multicolumn{1}{|c|}{Allowed Response:} \\
		\begin{minipage}[c]{.89\textwidth}
			\begin{tabular}{|p{\textwidth}|}
				\hline
				\\[-0.8em]
				\multicolumn{1}{|c|}{\textbf{\response{Choice}{$\L$}}} \\
				\textit{Asks the solver to perform the choices in list $\L$. $\L$ contains tuples of the form ($at,sign$), where $at \in \A'$ and $sign \in \{p,n,f\}$. Given tuple $(at,sign)$, the solver chooses $at$ if $sign=p$, $\neg at$ if $sign=n$, and if $sign=f$ the CDCL is left free to choose the sign.}\\
				\hline
				\\[-0.8em]
				\multicolumn{1}{|c|}{\textbf{\response{Unroll}{$\ell$}}} \\
				\textit{Asks the solver to backtrack until $\ell$ becomes undefined. To trigger a restart set $\ell=\bot$.}\\
				\hline
				\\[-0.8em]
				\multicolumn{1}{|c|}{\textbf{\response{Fallback}{$n, \I, \F, \S$}}} \\
				\textit{Asks the solver to use the default heuristics for the next $n$ choices.
					The default heuristic is enable permanently if $n \leq 0$.
					Parameters $\I, \F, \S$ initialize the default heuristics as follows:
					$\I: A \rightarrow \mathbb{N}$ provides the initial activity of atoms;
					$\F: A \rightarrow \mathbb{N}$ associates to atoms an amplifying factor;
					$\S: A \rightarrow \{p,n\}$ to provide a priority on the sign of literals.}\\ 
				\hline
				\\[-0.8em]
				\multicolumn{1}{|c|}{\textbf{\response{AddClause}{$\varphi$}}} \\
				\textit{Asks the solver to add the clause $\varphi$ to the formula.}\\
				\hline
			\end{tabular}	
		\end{minipage}\vspace*{2pt}\\ 
		\hline
	\end{tabular}
	\caption{Requests and responses.}
	\label{tab:responses}
\end{table}

The \request{GetChoice}{$I$} request is the main one, and is made 
by the solver when a non-deterministic choice is required.
A number of responses are possible besides the obvious one where 
the driver provides the solver with an ordered list of branching literals to be chosen.
The list can hold just one element as in many well-known heuristics, or can provide in one call several choices so to reduce communication overheads in domain-heuristics that may provide at once an entire plan of choices.
Alternative responses were inspired by domain-specific criteria.
For example, the driver is likely to recognize specific paths in the search space that cannot lead to a solution, so it can issue a \request{Unroll}{$\ell$} to cause a backtrack up to the last choice made before $\ell$.
Moreover, the driver could add some additional domain knowledge modeled by a clause $\varphi$ in the solver by answering \response{AddClause}{$\varphi$}.
Finally, there are cases (see next section) in which it makes sense to blend custom heuristics with the default heuristic, to exploit the strengths of both. To this end the driver can respond \response{Fallback}{$n,\I,\F,\S$} so to force the solver to use its default heuristic for the next $n$ choices. The parameters of \response{Fallback}{.} allow to customize this interaction by (possibly) tuning the activity values and the multiplying factors used by the default heuristic. 

In the following we provide a number of examples that show how to encode several well-known conflict-based and domain-specific heuristics. This way the reader can appreciate the generality of our proposal that cover the needs of a wide range of conflict-based heuristics as well as domain-heuristics proposed in the literature.
Nevertheless, we remark that our framework can be easily extended by adding new events/requests to fit the needs of future applications.

\begin{example}[Minisat heuristic]\label{ex:minisat}
	We first consider the \minisat heuristic as described in~\cite{DBLP:conf/sat/EenS03}.
	This criterion can be implemented in our framework by exploiting the events \event{StartSearch}{$\Gamma',\A'$} and \event{LearnClause}{$\varphi$}, as well as the \request{GetChoice}{$I$} request.
	The first is used to initialize the activity values of atoms in the input; the second 
	causes an increment the activity of atoms in the learned clause $\varphi$. 
	Finally, the request \request{GetChoice}{$I$} is answered with \response{Choice}{$\{(a,n)\}$} by the driver, where $a$ is the undefined atom in $I$ with the highest value of activity.
\end{example}

\begin{example}[Pigeonhole problem heuristic]\label{ex:pigeonhole}
	Consider the famous pigeonhole problem: given $n$ pigeons and $m$ holes, assign each pigeon to exactly one hole in such a way that two different pigeons do not share the same hole.
	It can be verified that state-of-the-art SAT solvers are not able to solve this problem efficiently when $n > m$, despite it is known that no solution exists in that case.
	A domain-specific heuristic based on this observation only uses the requests and no event.
	To \request{GetAtomsToBeFrozen}{$\A$} the driver answers to not remove any atom (providing \response{Freeze}{$\A$}), and  computes $n$ and $m$ by interpreting the atoms in $\A$.
	Once the first choice is requested, if $n > m$ the heuristic replies with \response{AddClause}{$\{\bot\}$} causing the termination of the CDCL returning \inco;
	otherwise, \response{fallback}{$0, \emptyset, \emptyset, \emptyset$} is sent to go on with the default heuristic.
\end{example}

\begin{example}[Rintanen's heuristic]\label{ex:rintanen}
	In \cite{DBLP:journals/ai/Rintanen12}, a heuristic for solving the planning problem modeled as SAT problem is proposed that intuitively forces the solver to prefer short and simple plans.
	This criterion can be implemented in our framework by using the event \event{Search}{$\Gamma',\A$}, and the request \request{GetChoice}{$I$}. 
	In fact, \event{Search}{.} allows the driver to identify \textit{actions} and \textit{goals} literals; while the answer to \request{GetChoice}{.} is computed by running the algorithm in Figure~3 of \cite{DBLP:journals/ai/Rintanen12}.
\end{example}

We mention that, one can easily obtain all the \vsids-like heuristics described in~\cite{DBLP:conf/sat/BiereF15} by applying small modifications to the solution presented in Example~\ref{ex:minisat}.

\section{Case study: Answer Set Programming}
In this section we report on a case study that provides pragmatical evidence of the applicability of our framework.
The case study is developed in the context of Answer Set Programming (ASP)~\cite{DBLP:journals/cacm/BrewkaET11}.
ASP is a declarative problem solving paradigm proposed in the area of logic programming and non-monotonic reasoning.
ASP is an ideal test bed for our framework for two main reasons: $(i)$ ASP solvers are nowadays efficient CDCL implementations; $(ii)$ ASP has been applied to a variety of complex problems including hard industrial ones.
In the following we first describe the implementation of our framework in a well-known ASP solver and, then, we report on the solution of two real-world configuration problems occurring in practice of Siemens: the Partner Units Problem (\pup) and the Combined Configuration Problem (\ccp).
These appeared to be among the hardest industrial benchmarks from the ASP Competition 2015~\cite{DBLP:conf/lpnmr/GebserMR15,aspweb}; indeed, on hard problem instances state-of-the-art ASP solvers were not effective.
Given the practical importance of these two problems, many researchers studied them in detail~\cite{Falkner2010,DBLP:conf/cpaior/AschingerDFGJRT11,DBLP:conf/ijcai/AschingerDGJT11,DBLP:conf/iaai/TeppanFF12,DBLP:conf/ictai/Drescher12,DBLP:conf/lpnmr/GebserRS15}. 
Their investigations resulted in a number of domain-specific heuristic approaches that, in some cases, significantly outperformed existing ASP solvers.
We experimented with these heuristics, thus applying our framework in practice,
to see whether these can be integrated successfully in a CDCL solver.

\subsection{Implementation of the framework}
The framework has been implemented as an extension of the ASP solver \wasp.
\wasp implements a variant of CDCL devised for evaluating ASP programs, thus featuring additional inference rules required for the evaluation of ASP programs, such as \emph{unfounded-based} and \emph{aggregate-based} inferences~\cite{DBLP:journals/ai/GebserKS12}.
The framework for specifying heuristics has been implemented in \textit{C++} and it offers an infrastructure for easy specification and testing of heuristic strategies. The user can adopt several languages for implementing external heuristics that range from imperative to declarative ones.
In particular, the heuristic strategies can be implemented in \textit{perl} and \textit{python} for obtaining fast prototypes and in \textit{C++} for obtaining better performances.
(In our tests, we observed that \textit{C++} implementations are often faster than equivalent \textit{perl} and \textit{python} counterparts, which are usually easier to develop instead).
In case the user would like to use a declarative language to implement a heuristic, our implementation provides a \textit{predicates-based} interface extending the one proposed in~\cite{DBLP:conf/aaai/GebserKROSW13}.
We next sketch the idea by assuming that (on the lines of in~\cite{DBLP:conf/aaai/GebserKROSW13}) one encodes the heuristic strategy in ASP by providing a logic program $\Pi$.
Each request of the CDCL algorithm causes an evaluation of $\Pi$ together with \textit{facts} modeling the request (intuitively, a fact is an atom that is always true).
The evaluation of $\Pi$ with the aforementioned facts produces a set of atoms (i.e., the answer set produced by a solver) interpreted as responses for the request.
For instance, consider Example~\ref{ex:pigeonhole}, the request \request{GetFrozenAtoms}{$\A$} is represented by the facts $get\_frozen\_atoms()$ and $atom(p(i,j))$ for each $p_{i,j} \in \A$.
A call to an external solver produces an answer set containing $freeze(p(i,j))$, for each $p_{i,j} \in \A$, that is interpreted as \response{Freeze}{$\A$}.
Other request and corresponding responses are handled in a similar manner.
Our implementation currently supports any external solver compliant with ASP solver output. 
The predicate-based interface can easily support variants of ASP (e.g., with \emph{preferences}~\cite{DBLP:conf/aaai/BrewkaD0S15}, \emph{CASP}~\cite{DBLP:conf/iclp/GebserOS09}, etc.) an could be adapted, in principle, also to support other languages based on predicate logic.

\subsection{Partner Units Problem}

\begin{figure}[b]
	\centering

\begin{tikzpicture}[x=3em,y=3em,baseline=0pt]
  \tikzstyle{track} =   [semithick, double, rounded corners=0.4em]
  \tikzstyle{sensor} =  [circle, fill, inner sep=0, minimum size=0.25em]
  \tikzstyle{zone} =    [black!60, densely dotted, rounded corners=0.2em]
  
  \draw[track] (0,0) -- ++(6,0)
               (0,0) -- ++(1,0) -- ++(1,1) -- ++(2,0) -- ++(1,-1) -- ++(1,0);
  
  \node[sensor, label=below:$s_1$] at (0.5,-0.15) {};
  \node[sensor, label=above:$s_2$] at (1.45,0.65) {};
  \node[sensor, label=below:$s_3$] at (1.5,-0.15) {};
  \node[sensor, label=above:$s_4$] at (4.55,0.65) {};
  \node[sensor, label=below:$s_5$] at (4.5,-0.15) {};
  \node[sensor, label=below:$s_6$] at (5.5,-0.15) {};
  
  \draw[zone] (-0.25,-0.75) rectangle node[left=.5em, label=left:$z_{1}$]{} ++(1,0.85);
  \draw[zone] (0.25,-0.85) rectangle node[below=2.7em, label=below:$z_{123}$]{} ++(1.5,2.25);
  \draw[zone] (1.2,0.5) rectangle node[above=.5em, label=above:$z_{24}$]{} ++(3.6,0.75);
  \draw[zone] (1.2,-0.75) rectangle node[below=.5em, label=below:$z_{35}$]{} ++(3.6,0.85);
  \draw[zone] (4.25,-0.85) rectangle node[below=2.7em, label=below:$z_{456}$]{} ++(1.5,2.25);
  \draw[zone] (5.25,-0.75) rectangle node[right=.5em, label=right:$z_{6}$]{} ++(1,0.85);

\end{tikzpicture}
\qquad
\begin{tikzpicture}[x=2em,y=2em,baseline=0pt]
  \tikzstyle{sensor} = [draw, circle, inner sep=0em]
  \tikzstyle{zone} =   [draw, rectangle, inner sep=.2em, minimum width=2em, rounded corners=0.2em]
  \tikzstyle{unit} =   [draw, rectangle]
  \tikzstyle{unit2zs} = [black!60, densely dotted, rounded corners=0.2em]
 
  \node[sensor] (s1) at (3.25,1.875) {$s_1$};
  \node[sensor] (s3) at (3.25,1.125) {$s_3$};
  \node[sensor] (s5) at (3.25,0.375) {$s_5$};
  \node[sensor] (s4) at (3.25,-0.375) {$s_4$};
  \node[sensor] (s6) at (3.25,-1.125) {$s_6$};
  \node[sensor] (s2) at (3.25,-1.875) {$s_2$};
  
  \node[zone] (z1)   at (0.75,1.875) {$z_{1}$};
  \node[zone] (z35)  at (0.75,1.125) {$z_{35}$};
  \node[zone] (z123) at (0.75,0.375) {$z_{123}$};
  \node[zone] (z456) at (0.75,-0.375) {$z_{456}$};
  \node[zone] (z6)   at (0.75,-1.125) {$z_{6}$};
  \node[zone] (z24)  at (0.75,-1.875) {$z_{24}$};
  
  \node[unit] (u1) at (2.15, 1.5) {$u_1$};
  \node[unit] (u2) at (2.15, 0) {$u_2$};
  \node[unit] (u3) at (2.15, -1.5) {$u_3$};
  
  \draw (z1.east)   -- (u1) -- (s1)
        (z35.east)  -- (u1) -- (s3)
        (z456.east) -- (u2) -- (s4)
        (z123.east) -- (u2) -- (s5)
        (z6.east)   -- (u3) -- (s6)
        (z24.east)  -- (u3) -- (s2);
  
  \draw (u1) -- (u2) -- (u3);

\end{tikzpicture}    
	\caption{A sample railway station layout and a solution of a corresponding \pup instance with $\ucap =\iucap=2$}  \label{fig:pup_layout}
\end{figure}
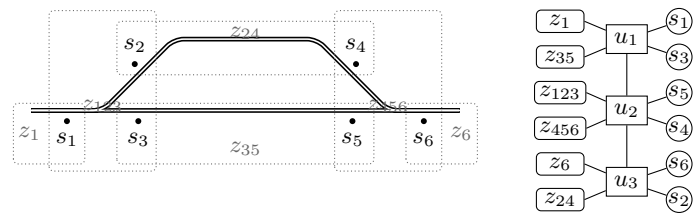

The \pup comes from the railway safety domain~\cite{Falkner2010}, but has a variety of other applications including security monitoring systems, peer-to-peer networks, etc.~\cite{DBLP:conf/iaai/TeppanFF12}. 
In order to ensure the safety of train movements, railway tracks (see Figure~\ref{fig:pup_layout}) are equipped with hardware \emph{sensors} $s_1,\dots,s_6$ which register when a wagon passes by. 
All registered events are forwarded from sensors to \emph{control units} $u_1,\dots,u_3$ belonging to a safety system. The latter has to prevent wagons of different trains from entering the same safety \emph{zone} of the tracks $z_1,\dots,z_{24}$, hence avoiding unwanted collisions. 
To solve the problem one needs to (1) assign every zone and sensor to a control unit such that every unit is connected to at most \ucap sensors and at most \ucap zones. Moreover, (2) if a sensor contributes to a zone, it must be placed on the same or on a connected unit (partner unit). However, every unit can be a partner of at most \iucap other units.

\paragraph{Heuristics.} For our evaluation we selected the \qpup heuristic and its derivatives, since search algorithms using these heuristics were able to outperform best ASP solvers.

\begin{itemize}
	\item \qpup~\cite{DBLP:conf/iaai/TeppanFF12} generates an order of zones and sensors in the first step by selecting some zone as a root and traversing the input zone-sensor relations breadth-first. 
	Then, a depth-first search selects the next unassigned zone or sensor in the order and assigns it to a new unit. If the assignment cannot be done due to a violation of requirement (2), then the already used units are tested starting from the last created one. 
	In case \qpup is not able to assign a zone or sensor to any unit it unrolls the last made assignment and tries another possibility. 
	Finally, if no more assignments are possible, \qpup falls back to the default heuristic.
	
	\item \qspup is a modification of \qpup that tests already used units first and a new one afterwards.
	
	\item \pred~\cite{DBLP:conf/ictai/Drescher12} tries first to connect an unassigned zone or sensor to one of the units that already have connections to neighbour zones or sensors, i.e.\ reachable over at most two input zone-sensor relations. If this fails, \pred tries a new unit first and used ones subsequently.
\end{itemize}

\paragraph{Implementation details.} We implemented the four heuristics in \wasp using the C++ interface given in Section~\ref{sec:heur}. A heuristic is initialized as \event{Search}{$\Gamma',\A'$} event is invoked by the solver. Next, the ordering of vertices is generated and used to determine sets of chosen literals $L$ in response to \request{GetChoice}{$I$} requests. In addition, a  heuristic registers \event{UnrollLit}{$\ell$} events in order to synchronize its inner state with decisions made by \wasp between two subsequent \request{GetChoice}{$I$} requests. If a heuristic finds that the current partial interpretation is inconsistent, i.e.\ no further assignments can be made, it backtracks and sends \response{Unroll}{$\ell$} to \wasp. In this case, $\ell$ is one of the literals that were selected by a heuristic at the previous choice event. Finally, the fallback from \qpup-based heuristics to the default one was implemented using \response{Fallback}{$0,\emptyset,\emptyset,\emptyset$} response. 

\paragraph{Setup.} The evaluation was done by using the set of instances provided by Siemens\footnote{http://demo2-iwas.uni-klu.ac.at/pupsolver/benchmark.rar}. This package comprises instances of four types: (1) double, (2) double-variant, (3) triple and (4) grid. The first three types are instances representing  topologies occurring frequently in practice, whereas grids are parts of real railway systems (see~\cite{DBLP:conf/cpaior/AschingerDFGJRT11} for details).
The instances were tested with two encodings: 
\pup-E1 -- the original straightforward encoding published in~\cite{DBLP:conf/cpaior/AschingerDFGJRT11} (Section 3.1, p.\ 6) and \pup-E2 -- a complex encoding comprising symmetry breaking and ordering rules used in the ASP Competition 2015.
Moreover, in all experiments with \wasp using heuristics, we removed rule 9 (line 14) from \pup-E2. This rule is a symmetry breaker that does not allow our heuristics to select the starting zone.
For each encoding we grounded all instances with \gringo (version 4.5.3). 
Then each solver was run for 900 sec.\  on a system equipped with i7-3030K CPU, 64GB RAM and Ubuntu 11.10.

\begin{figure}[tb]
	\centering
\begin{tikzpicture}[scale=1]
\begin{axis}[
  xlabel={\# instances solved out of 36}
, ylabel={runtime $[sec]$}
, xmin=0, xmax=37
, ymin=0
, xtick={0,6,12,18,24,30,36}
, legend style={at={(0.3,.96)},anchor=north, font=\footnotesize, draw=none}
, legend columns=1
, cycle list name=cactus
, grid=both
]
\addplot +[unbounded coords=jump] table[col sep=semicolon, y index=3] {./pup.csv};
\addlegendentry{\clasp}
\addplot +[unbounded coords=jump] table[col sep=semicolon, y index=5] {./pup.csv};
\addlegendentry{\wasp \qpup}
\addplot +[unbounded coords=jump] table[col sep=semicolon, y index=7] {./pup.csv};
\addlegendentry{\wasp \qspup}
\addplot +[unbounded coords=jump] table[col sep=semicolon, y index=8] {./pup.csv};
\addlegendentry{\wasp \pred}
\end{axis}
\end{tikzpicture}
	\caption{Experimental results for 36 \pup  instances~\cite{paperwebsite} with \pup-E1 encoding\label{fig:eval:a}}
\end{figure}
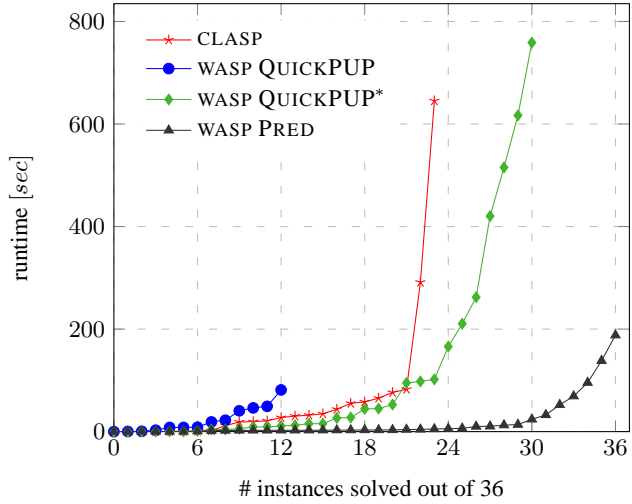

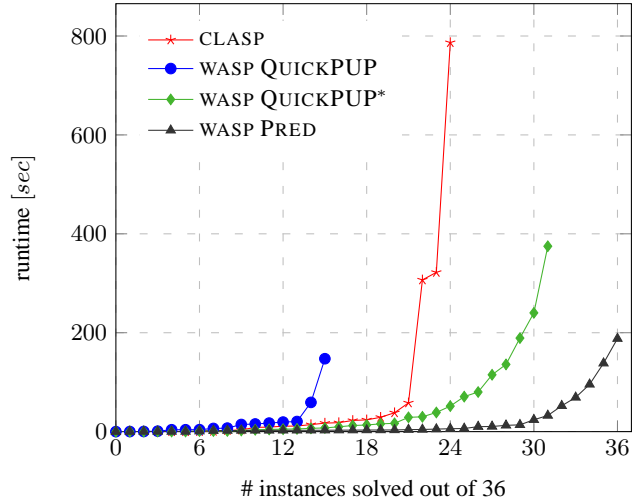
\begin{figure}[tb]
	\centering
\begin{tikzpicture}[scale=1]
\begin{axis}[
  xlabel={\# instances solved out of 36}
, ylabel={runtime $[sec]$}
, xmin=0, xmax=37
, ymin=0
, xtick={0,6,12,18,24,30,36}
, legend style={at={(0.3,.96)},anchor=north, font=\footnotesize, draw=none}
, legend columns=1
, cycle list name=cactus
, grid=both
]
\addplot +[unbounded coords=jump] table[col sep=semicolon, y index=1] {./pup.csv};
\addlegendentry{\clasp}
\addplot +[unbounded coords=jump] table[col sep=semicolon, y index=4] {./pup.csv};
\addlegendentry{\wasp \qpup}
\addplot +[unbounded coords=jump] table[col sep=semicolon, y index=6] {./pup.csv};
\addlegendentry{\wasp \qspup}
\addplot +[unbounded coords=jump] table[col sep=semicolon, y index=8] {./pup.csv};
\addlegendentry{\wasp \pred}
\end{axis}
\end{tikzpicture}
	\caption{Experimental results for 36 \pup instances~\cite{paperwebsite} with \pup-E2 encoding\label{fig:eval:b}}
\end{figure}

\paragraph{Results.} A summary of our evaluation results is presented in Figures~\ref{fig:eval:a} and \ref{fig:eval:b}. The results of \clasp~\cite{DBLP:conf/aaai/GebserKROSW13}, which was used as a reference system, show that it was able to solve at most 23 instances using the tested encodings. Moreover, in both experiments \clasp failed to find a model for most of the double as well as for some double-variant and triple instances. 
Also \wasp with \qpup solved only 12 and 15 instances using \pup-E1 and \pup-E2 respectively. The main reason for this discrepancy between our results and the ones obtained by~\cite{DBLP:conf/iaai/TeppanFF12} is due to inability of the underlying ASP solver to generate new units on-the-fly. Therefore, the number of ground literals denoting available units is determined in the grounding step and cannot be changed later. \cite{DBLP:conf/iaai/TeppanFF12}, instead, try to quickly build a solution by using new units whenever possible. This strategy appears to be successful for greedy algorithms, but works poorly as a heuristic for \wasp. 

The two remaining heuristics assign zones and sensors to already existing units first. 
This allowed \wasp to outperform \clasp and \qpup in all experiments. 
The best result was observed for \wasp with \pred, which  solved \emph{all} instances independently of the encoding. Moreover, \pup-E2 allowed \wasp to reduce its average solving time to only 27 sec. 
Finally, we would like to emphasize the fact that \pred allowed \wasp to find solutions for all instances with the \pup-E1 encoding. 
This is an important feature of our approach. 
It allows a programmer to easily model a problem using a declarative language, like ASP-Core, and then to scale performance of the solver using heuristics defined in a procedural way. 

\subsection{Combined Configuration Problem}

Similarly to the \pup, this problem was derived from a number of real-world problems including railway interlocking systems, safety automation, resource distribution, etc. The \ccp is an abstract problem that demonstrates an important case occurring in practice of Siemens when a complex problem can be represented as a set of subproblems~\cite{DBLP:conf/lpnmr/GebserRS15}.

\begin{figure}[bt]	
	\centering
	\tikzstyle{node} = [circle, draw, text centered, inner sep=0em, minimum width=15pt ]
\tikzstyle{rnode} = [node, fill=red!20]
\tikzstyle{bnode} = [node, fill=blue!20]
\tikzstyle{gnode} = [node, fill=green!20]
\tikzstyle{edge}  = [line width=1.3pt, ->]
\tikzstyle{path1} = [edge,red]
\tikzstyle{path2} = [edge,darkgreen]
\tikzstyle{a1}    = [dashed]
\tikzstyle{a2}    = [a1,draw=gray,very thick]

\tikzset{
    bin/.style={
        label={[label distance=-5pt]-(-60):\scriptsize#1}}
}

\begin{tikzpicture}[semithick,auto]
\node at (-1.6,0)    [node](b1) {$b_{1}$};    
\node at (-0.8,0)    [node](p1) {$p_{1}$};    
\node at (0,0)     [node](b2) {$b_{2}$};
\node at (.8,0)     [node](p3) {$p_{3}$};
\node at (1.6,0)     [node](b5) {$b_{5}$};

\node at (-.8,-.9)    [node](b3) {$b_{3}$};    
\node at (0,-.9)     [node](e1) {$p_{2}$};
\node at (.8,-.9)     [node](b4) {$b_{4}$};

\path[path1] (b1)  edge  node {} (p1);
\path[edge]  (p1)  edge  node {} (b2);
\path[edge]  (b2)  edge  node {} (p3);
\path[path2] (p3)  edge  node {} (b5);
\path[path1] (p1)  edge  node {} (b3);
\path[edge]  (b3)  edge  node {} (e1);
\path[edge]  (e1)  edge  node {} (b4);
\path[path2] (b4)  edge  node {} (p3);
    
\draw [rounded corners=10pt,dashed,darkgray,thick] (-1.9,0.5)--(-2.1,-0.2) -- node[below,fill=white] {\textcolor{black}{$\mathit{BE}_1$}}  (-1.2,-1.3) -- (-.45,-1.3) -- (-0.5,-0.4) -- (0.35,-0.45) -- (.35,0.45)--cycle ;

\draw [rounded corners=10pt,dashed,gray,very thick] (1.9,0.6)--(2.1,-0.2) -- node[below,fill=white] {\textcolor{black}{$\mathit{BE}_2$}}  (1.2,-1.3) -- (.35,-1.3) -- (0.35,-0.45) -- (-0.45,-0.6) -- (-.45,0.6)--cycle ;

\end{tikzpicture}
\qquad
\begin{tikzpicture}[semithick,auto]
\node at (-1.6,0)  [rnode,a1,bin={2}]   (b1) {$b_{1}$};    
\node at (-0.8,0)  [rnode,bin={1}]      (p1) {$p_{1}$};    
\node at (0,0)     [rnode,a1,bin={2}]   (b2) {$b_{2}$};
\node at (.8,0)    [gnode,bin={1}]      (p3) {$p_{3}$};
\node at (1.6,0)   [gnode,a2,bin={2}]   (b5) {$b_{5}$};

\node at (-.8,-.9) [rnode,a1,bin={2}]   (b3) {$b_{3}$};    
\node at (0,-.9)   [bnode,bin={1}]      (e1) {$p_{2}$};
\node at (.8,-.9)  [gnode,a2,bin={2}]   (b4) {$b_{4}$};

\path[path1] (b1)  edge  node {} (p1);
\path[edge]  (p1)  edge  node {} (b2);
\path[edge]  (b2)  edge  node {} (p3);
\path[path2] (p3)  edge  node {} (b5);
\path[path1] (p1)  edge  node {} (b3);
\path[edge]  (b3)  edge  node {} (e1);
\path[edge]  (e1)  edge  node {} (b4);
\path[path2] (b4)  edge  node {} (p3);

\end{tikzpicture}
	\caption{Sample \ccp instance (left) and its solution (right).}	
	\label{fig:ccp}
\end{figure}
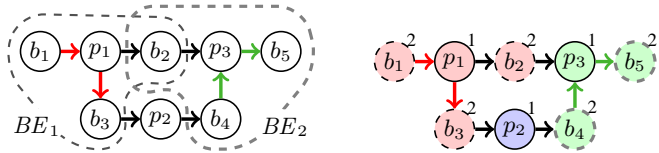

A \ccp instance is defined by a directed acyclic graph $G=(V,E)$ representing a track layout (see Figure~\ref{fig:ccp}). Each vertex has a type, e.g.\ $b$ and $p$, with an associated size, e.g. 1 and 3. Furthermore, an instance includes two disjoint paths $P_1$ and $P_2$, shown in red and green, as well as several sets of border elements, e.g.\ $\mathit{BE}_1=\{b_1,b_2,b_3\}$ and $\mathit{BE}_2=\{b_2,b_4,b_5\}$. Finally, it defines parameters specifying a maximal number of border elements per safety area $M$, a number of colors $C$, a number of available bins $B$ per color and capacity of a bin $K$.
The goal is to find a solution that satisfies all requirements of the following subproblems:
\begin{description}
	\item[(P1) Coloring] Every vertex must have exactly one color.
	
	\item[(P2) Bin-Packing] For every set $V_i \subseteq V$ comprising all vertices of one color, assign each vertex $v_j \in V_i$ to exactly one bin such that for each bin the sum of the vertex sizes is less or equal to $K$ and at most $B$ bins are used.
	
	\item[(P3) Disjoint Paths] Colors of any two vertices $v_i \in P_1$ and $v_j \in P_2$ must be different.
	
	\item[(P4) Matching] For every input set of border elements $\mathit{BE}_i$ find a safety area $A_i\subseteq \mathit{BE}_i$ such that all border elements of one area are colored in one color, $|A_i| \leq M$ and every border element is exactly in one area.
	
	\item[(P5) Connectedness] Two vertices of the same color must be connected via a path that comprises only vertices of this color.
\end{description}

\begin{example}
	Let the sample \ccp instance comprise the following parameters: $B=2$, $K=3$, $C=3$ and $M=3$. The solution presented in Figure~\ref{fig:ccp} assigns 3 colors to vertices, places them into 2 bins, shown in subscript of each vertex, and defines two safety areas $A_1=\{b_1,b_2,b_3\}$ and $A_2=\{b_4,b_5\}$.
\end{example}

\paragraph{Heuristics.}
The \ccp has a number of heuristics suggested in~\cite{DBLP:conf/lpnmr/GebserRS15} which can be used to obtain a partial solution of a problem instance. It turned out that it is quite simple to find a heuristic that solves one or some of the subproblems. However, no complete heuristics, like \qpup or \pred for \pup, are known for \ccp. 

The heuristics suggested in~\cite{DBLP:conf/lpnmr/GebserRS15} were designed to be used with the heuristic interface of \clasp. 
This interface allows a programmer to define a heuristic function in the program using $\_heuristic/4$ predicate. If a ground ``heuristic'' atom over this predicate becomes true in a partial interpretation, then terms of the atom are used by \clasp to modify behavior of the \vsids heuristic. 
Therefore, given a \ccp instance, the proposed approach first runs a heuristic search method to find a partial solution of the instance. Next, \clasp is forced to start the search from the found partial solution by expending the \ccp instance with facts comprising ``heuristic'' atoms representing it. 
Gebser et al. \cite{DBLP:conf/lpnmr/GebserRS15} use three heuristics:
\begin{itemize}
	\item  \accp{1} solves the matching problem (P4). In every iteration it assigns an unassigned border element to a safety area which comprises the minimum number of already assigned elements. 
	
	\item  \accp{2}. This heuristic is designed to solve three subproblems (P1), (P2) and (P5). First, it sets the color to 1 and adds some vertex $v \in V$ to a queue $Q$. Next, it retrieves an element $v_i \in Q$, assigns it the current color, places it to a bin according to the best-fit heuristic and adds all uncolored neighbors of $v_i$ to $Q$. The process continues, until no further vertices can be placed to bins. In this case the algorithm increases the value of current color and adds some uncolored node $v_j$ to $Q$.
	
	\item \mccp. This is a combination of the previous two which first applies \accp{1} and then uses \accp{2} to extend the output of \accp{1}.
\end{itemize}

All original heuristics were designed to run only once prior to solving whereas in our solver a heuristic is queried multiple times. Therefore, we suggest the following modifications: 
\begin{itemize}
	\item \bccp is based on \accp{2} which is extended with a fallback to a default solver heuristic if (i) a partial solution is found or (ii) the 10 seconds timeout is reached or (iii) \accp{2} cannot make a choice. The latter situation might occur if all atoms used in subproblems (P1), (P2) and (P5) already have assigned truth values.
	
	\item \bfccp. Originally \bccp processes the vertices of $G$ in no specific order. To increase its performance we extended \bccp with an ordering that assigns a score of $2$ to all vertices with only incoming or outgoing edges, a score of $1$, it is in a path, and $3$, if both applies. Then, \bfccp runs \bccp such that in every iteration it sorts the unprocessed vertices in the decreasing order of their scores.
	
	\item \bfaccp. This version of \bfccp uses a fallback strategy which alternates between \bfccp and a default solver heuristic. Thus, \bfaccp runs \bfccp with the first ordering until it cannot make any further choices. Then, it switches to the default heuristic for the next 10 seconds. If solver was unable to get a complete solution within this time, \bfaccp creates a new ordering and executes \bfccp once more. In the case all possible orderings starting with vertices with a score higher than zero has been used, \bfaccp falls back to the default heuristic permanently. The idea is to allow the default heuristic of the solver to acquire enough representative conflicts and hence to improve its performance.
\end{itemize}

\paragraph{Implementation details.} The heuristics were implemented similarly to the ones used for the \pup using the C++ version of the \wasp interface. When \bfaccp decides to use the default solver heuristic, it starts a timer and responses all \request{GetChoice}{$I$} requests with \response{Fallback}{$1,\emptyset,\emptyset,\emptyset$}. When the timeout is reached, \bfaccp generates another ordering, restarts the search using \response{Unroll}{$\bot$} and answers all subsequent requests itself.

\paragraph{Setup.} We evaluated \bccp, \bfccp and \bfaccp on the set of hard instances\footnote{http://isbi.aau.at/hint/images/ccp/ccp.tgz} and the encoding submitted to the ASP Competition 2015. 
The package contains a set of real-world instances used in the evaluation of~\cite{DBLP:conf/lpnmr/GebserRS15} as well as a number of new instances. The latter have a grid-like graph structure in which all vertices are of the same type.
The evaluation was done in the same settings as the \pup and similar to~\cite{DBLP:conf/lpnmr/GebserRS15}. 
The average grounding time was about 2 seconds and is excluded from results.

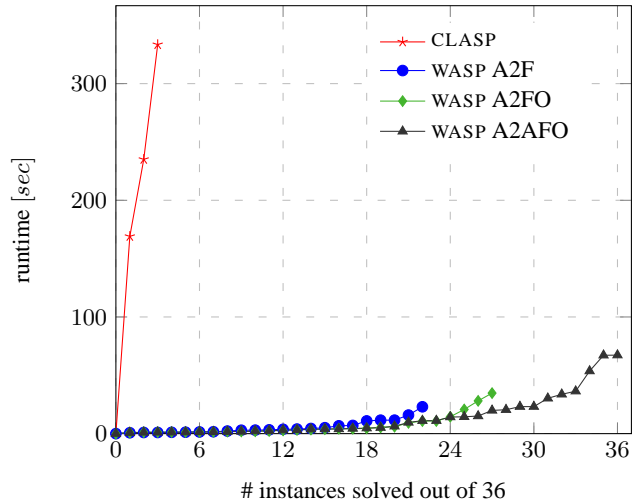
\begin{figure}[tb]
	\centering
\begin{tikzpicture}[scale=1]
\begin{axis}[
  xlabel={\# instances solved out of 36}
, ylabel={runtime $[sec]$}
, xmin=0, xmax=37
, ymin=0
, xtick={0,6,12,18,24,30,36}
, legend style={at={(0.7,.96)},anchor=north, font=\footnotesize, draw=none}
, legend columns=1
, cycle list name=cactus
, grid=both
]
\addplot +[unbounded coords=jump] table[col sep=semicolon, y index=1] {./ccp.csv}; 
\addlegendentry{\clasp}


\addplot +[unbounded coords=jump] table[col sep=semicolon, y index=4] {./ccp.csv}; 
\addlegendentry{\wasp \bccp}

\addplot +[unbounded coords=jump] table[col sep=semicolon, y index=5] {./ccp.csv}; 
\addlegendentry{\wasp \bfccp}

\addplot +[unbounded coords=jump] table[col sep=semicolon, y index=6] {./ccp.csv};
\addlegendentry{\wasp \bfaccp}
\end{axis}
\end{tikzpicture}
\vspace{-10pt}
	\vspace{5pt}
	\caption{Evaluation results for 36 CCP instances with~\cite{DBLP:conf/lpnmr/GebserRS15} encoding \label{fig:eval:c}}
\end{figure}

\paragraph{Results.} The evaluation results, presented in Figure~\ref{fig:eval:c}, show that the reference system could solve 3 out of 36 instance with an average solution time of 246 sec.  
Our approach using \bccp, \bfccp and \bfaccp heuristics was able to solve 22, 27 and 36 instances with an average solution time 6, 7, and 14 sec.\  resp. \bfaccp heuristic has a greater solving time only because it was able to solve the hardest instances. Our approach shows also better results in comparison with ~\cite{DBLP:conf/lpnmr/GebserRS15}, which solved 6 out of 16 real-world instances.

Moreover, since \bfaccp is based on an alternating usage of two heuristics we performed an additional evaluation. The goal was to test whether conflict learning is essential or the heuristic has to find a ``correct'' ordering as it was often the case for \pup heuristics. Therefore, we saved the last ordering \bfaccp used before the last fallback and then started a new experiment in which this ordering was used first. The results show that in none of the experiments this strategy allowed \wasp to find a solution with only one fallback. In all test cases \bfaccp run \accp{2} at least twice, i.e.\ generates at least two orderings, to find a solution. This indicates that learned clauses are essential.

\section{Conclusion}
The branching heuristic is a key ingredient of the CDCL algorithm.
Despite several effective heuristics have been proposed in the literature, 
their integration in existing solvers is usually nontrivial.
The paper proposed a novel framework that simplifies the design of new heuristics for CDCL solvers by distilling the key ingredients useful to drive the search.
The proposal has been validated in a case study where our framework was employed for solving effectively two hard industrial problems. Results are positive: hard instances that were not solvable by previous approaches are efficiently solved by our implementation.
Ongoing work concerns the implementation of our framework in other well-known solvers. Moreover, we plan to apply it in further practical use cases.

\section*{Acknowledgements}
This work was partially supported by the Carinthian Science Fund (KWF) contract KWF-3520/26767/38701, the Austrian Science Fund (FWF) contract I 2144 N-15, the Italian Ministry of University, Research under PON project ``Ba2Know (Business Analytics to Know) Service Innovation -- LAB'', No. \ PON03PE\_00001\_1, and by the Italian Ministry of Economic Development under project ``PIUCultura (Paradigmi Innovativi per l'Utilizzo della Cultura)'' n.\ F/020016/01--02/X27.


\begin{thebibliography}{10}
	\providecommand{\url}[1]{\texttt{#1}}
	\providecommand{\urlprefix}{URL }
	
	\bibitem{DBLP:conf/lpnmr/AlvianoDLR15}
	Alviano, M., Dodaro, C., Leone, N., Ricca, F: {Advances in WASP}.
	In {LPNMR} 2015. pp. 40--54. (2015)
	
	\bibitem{DBLP:conf/cpaior/AschingerDFGJRT11}
	Aschinger, M., Drescher, C., Friedrich, G., Gottlob, G., Jeavons, P., Ryabokon,
	A., Thorstensen, E.: {Optimization Methods for the Partner Units Problem}.
	In: {CPAIOR} 2011. pp. 4--19. (2011)
	
	\bibitem{DBLP:conf/ijcai/AschingerDGJT11}
	Aschinger, M., Drescher, C., Gottlob, G., Jeavons, P., Thorstensen, E.:
	{Tackling the Partner Units Configuration Problem}. In: {IJCAI} 2011. pp.
	497--503. (2011)
	
	\bibitem{DBLP:conf/ijcai/AudemardS09}
	Audemard, G., Simon, L.: {Predicting Learnt Clauses Quality in Modern {SAT} Solvers}. In: {IJCAI} 2009. pp. 399--404. (2009)
	
	\bibitem{DBLP:journals/jair/BeameKS04}
	Beame, P., Kautz, H.A., Sabharwal, A.: {Towards Understanding and Harnessing
		the Potential of Clause Learning}. J. Artif. Intell. Res. {(JAIR)}  22, 319--351 (2004)
	
	\bibitem{DBLP:conf/sat/BiereF15}
	Biere, A., Fr{\"{o}}hlich, A.: {Evaluating {CDCL} Variable Scoring Schemes}.
	In: {SAT} 2015. pp. 405--422. (2015)
	
	\bibitem{DBLP:series/faia/2009-185}
	Biere, A., Heule, M., van Maaren, H., Walsh, T. (eds.): {H}andbook of
	{S}atisfiability, Frontiers in Artificial Intelligence and Applications, vol. 185. {IOS} Press (2009)
	
	\bibitem{DBLP:conf/aaai/BrewkaD0S15}
	Brewka, G., Delgrande, J.P., Romero, J., Schaub, T.: {asprin: Customizing
		Answer Set Preferences without a Headache}. In: {AAAI} 2015. pp. 1467--1474. (2015)
	
	\bibitem{DBLP:journals/cacm/BrewkaET11}
	Brewka, G., Eiter, T., Truszczynski, M.: {Answer set programming at a glance}.	Commun. {ACM}  54(12),  92--103 (2011)
	
	\bibitem{DBLP:conf/ictai/Drescher12}
	Drescher, C.: {T}he {P}artner {U}nits {P}roblem a {C}onstraint {P}rogramming
	{C}ase {S}tudy. In: {ICTAI} 2012. pp. 170--177.	(2012)
	
	\bibitem{DBLP:conf/sat/EenB05}
	E{\'{e}}n, N., Biere, A.: {Effective Preprocessing in {SAT} Through Variable and Clause Elimination}. In: {SAT} 2005. pp. 61--75. (2005)
	
	\bibitem{DBLP:conf/sat/EenS03}
	E{\'{e}}n, N., S{\"{o}}rensson, N.: {An Extensible SAT-solver}. In: {SAT} 2003.	pp. 502--518. (2003)
	
	\bibitem{Falkner2010}
	Falkner, A., Haselb{\"{o}}ck, A., Schenner, G.: {Modeling Technical Product	Configuration Problems}. In: Workshop on Configuration, ECAI 2010. pp. 40--46.	(2010)
	
	\bibitem{InvitedFriedrich15}
	Friedrich, G.: Industrial success stories of {ASP} and {CP}: What's still open?	(2015), joint invited talk at ICLP and CP 2015 
	
	\bibitem{DBLP:conf/aaai/GebserKROSW13}
	Gebser, M., Kaufmann, B., Romero, J., Otero, R., Schaub, T., Wanko, P.:
	{Domain-Specific Heuristics in Answer Set Programming}. In: {AAAI} 2013.
	 (2013)
	
	\bibitem{DBLP:journals/ai/GebserKS12}
	Gebser, M., Kaufmann, B., Schaub, T.: {Conflict-driven answer set solving: From theory to practice}. Artif. Intell.  187,  52--89 (2012)
	
	\bibitem{DBLP:conf/lpnmr/GebserMR15}
	Gebser, M., Maratea, M., Ricca, F.: The design of the sixth answer set
	programming competition report. In: {LPNMR} 2015. pp. 531--544. (2015)
	
	\bibitem{aspweb}
	Gebser, M., Maratea, M., Ricca, F.: The sixth answer set programming
	competition web site. \url{http://aspcomp2015.dibris.unige.it/}
	
	\bibitem{DBLP:conf/iclp/GebserOS09}
	Gebser, M., Ostrowski, M., Schaub, T.: {Constraint Answer Set Solving}. In: {ICLP} 2009. pp. 235--249. (2009)
	
	\bibitem{DBLP:conf/lpnmr/GebserRS15}
	Gebser, M., Ryabokon, A., Schenner, G.: {Combining Heuristics for Configuration
		Problems Using Answer Set Programming}. In: {LPNMR} 2015.   pp. 384--397.   (2015) 
	
	\bibitem{DBLP:journals/ai/HutterXHL14}
	Hutter, F., Xu, L., Hoos, H.H., Leyton{-}Brown, K.: {Algorithm runtime
		prediction: Methods {\&} evaluation}. Artif. Intell.  206,  79--111 (2014) 
	
	\bibitem{DBLP:conf/dac/MoskewiczMZZM01}
	Moskewicz, M.W., Madigan, C.F., Zhao, Y., Zhang, L., Malik, S.: {Chaff:
		Engineering an Efficient {SAT} Solver}. In: {DAC} 2001. pp. 530--535.  (2001)
	
	\bibitem{DBLP:journals/jacm/NieuwenhuisOT06}
	Nieuwenhuis, R., Oliveras, A., Tinelli, C.: {Solving {SAT} and {SAT} Modulo Theories: From an abstract Davis--Putnam--Logemann--Loveland procedure to DPLL(\emph{T})}. J. {ACM}  53(6),  937--977 (2006) 
	
	\bibitem{DBLP:journals/ai/Rintanen12}
	Rintanen, J.: {Planning as satisfiability: Heuristics}. Artif. Intell.  193,
	45--86 (2012) 
	
	\bibitem{DBLP:journals/tc/Marques-SilvaS99}
	Silva, J.P.M., Sakallah, K.A.: {{GRASP:} {A} Search Algorithm for Propositional
		Satisfiability}. {IEEE} Trans. Computers  48(5),  506--521 (1999) 
	
	\bibitem{DBLP:conf/iaai/TeppanFF12}
	Teppan, E.C., Friedrich, G., Falkner, A.A.: {QuickPup: {A} Heuristic
		Backtracking Algorithm for the Partner Units Configuration Problem}. In: IAAI 2012. (2012) 
	
	\bibitem{paperwebsite}
	Heuristic Wasp Website. {\url{http://yarrick13.github.io/hwasp/}}
	
	\bibitem{DBLP:conf/iccad/ZhangMMM01}
	Zhang, L., Madigan, C.F., Moskewicz, M.W., Malik, S.: {Efficient Conflict
		Driven Learning in Boolean Satisfiability Solver}. In: {ICCAD} 2001. pp.
	279--285. (2001)
	
\end{thebibliography}
\end{document}